\documentclass[10pt,twocolumn,letterpaper]{article}

\usepackage{cvpr}              
\definecolor{cvprblue}{rgb}{0.21,0.49,0.74}
\usepackage[pagebackref,breaklinks,colorlinks,allcolors=cvprblue]{hyperref}
\usepackage{multirow}

\def\confName{CVPR}
\def\confYear{2026}

\title{CoT-Edit: Let CoT Guide Instruction Video Editing}

\author{
    Sen Liang$^{1}$\thanks{Equal contribution.} \quad
    Fengbin Guan$^{1}$\footnotemark[1] \quad
    Youliang Zhang$^{3}$ \quad
    Xin Li$^{1}$\thanks{Corresponding authors.} \quad  
    Zhibo Chen$^{1,2}$\footnotemark[2] \\
    $^{1}$University of Science and Technology of China \\
    $^{2}$Zhongguancun Academy \quad 
    $^{3}$Tsinghua University \\
    {\tt\small \{liangsen, guanfb\}@mail.ustc.edu.cn} \quad
    {\tt\small zhangyou24@mails.tsinghua.edu.cn} \\
    {\tt\small \{xin.li, chenzhibo\}@ustc.edu.cn}
}

\begin{document}
\maketitle

\begin{abstract}
Text-driven instruction-based video editing in complex scenes remains challenging: purely textual prompts often fail to capture precise spatial relationships and physical constraints, resulting in target ambiguity and physically implausible outcomes. 
To address this, we propose a plan--guide--edit framework that explicitly bridges semantic intent and spatial execution. 
In our framework, a Chain-of-Thought (CoT)-enhanced multimodal large language model (MLLM) serves as a planner, performing structured reasoning over the video and instructions to derive a precise sequence of bounding boxes and attribute-enriched editing directives.  
These spatial priors then guide a box-conditioned mask generator, transforming ambiguous global retrieval into localized, context-aware refinement and producing masks that more accurately capture object scale, contact relationships, and placement.
Building on these spatial and semantic signals, a diffusion-based editor integrates the masks, enriched instructions, and frame features to render high-fidelity edits that remain temporally coherent and spatially well aligned. 
Trained first in a modular manner and then jointly, our framework achieves superior performance with reduced data requirements, delivering precise localization in scenes with multiple similar objects and physically consistent object additions, and extensive experiments demonstrate state-of-the-art performance over multiple strong baseline methods. More details are available at: \url{https://github.com/flying-sky999/CoT-Edit}
\end{abstract}

\begin{figure}[t]
  \centering
  \includegraphics[width=\linewidth]{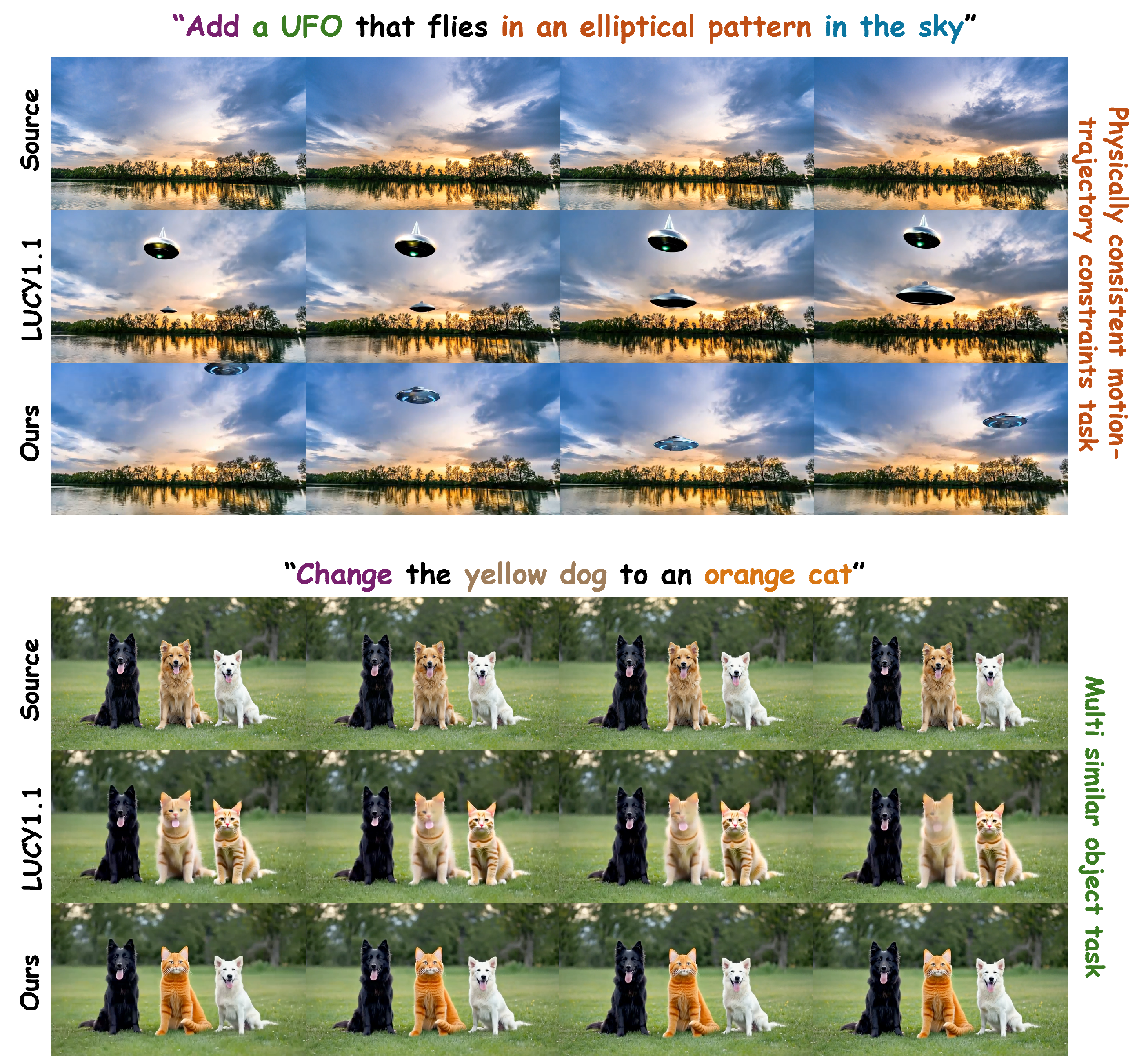}
  \caption{
    Illustration of the challenges in text-driven instruction-based video editing. 
    Given instructions such as ``Add a UFO that flies in an elliptical pattern in the sky'' (top) and ``Change the yellow dog into an orange cat'' (bottom), text-only methods struggle to respect trajectory constraints and to reliably select the intended target among multiple similar objects. 
  }
  \label{fig:motivation}
\end{figure}

\section{Introduction}
\label{sec:intro}

Instruction-based video editing~\cite{InstructX, InstructPix2Pix, InstructVid2Vid} has emerged as a transformative capability, enabling users to manipulate video content through natural language without requiring professional post-production skills, with applications spanning content creation and augmented reality. However, in complex real-world scenarios involving multiple coexisting objects, densely distributed similar entities, or dynamic interactions, reliance solely on textual prompts frequently leads to control failures. Purely text-driven editing models~\cite{Lucyedit, Insvie-1m, InstructX} must simultaneously comprehend cross-frame semantics, localize targets, and execute edits based on ambiguous linguistic signals. This multifaceted challenge compels models to depend on large-scale datasets and high-capacity models, yet these models still frequently result in localization drift, erroneous edits, and temporal instability due to insufficient spatial grounding and enforcement of physical constraints. As illustrated in Fig.~\ref{fig:motivation}, even seemingly simple instructions such as ``Add a UFO that flies in an elliptical pattern in the sky'' or ``Change the yellow dog into an orange cat'' can lead text-only methods to violate motion trajectories or to alter the wrong dog among several similar ones.

A feasible way to mitigate these issues is to introduce additional conditioning to decouple \textit{what to edit} from \textit{where to edit}. Masks are among the most practical spatial conditioning signals, offering clear cues that transform global retrieval into controllable local editing, reducing data requirements and enhancing controllability. However, if masks are generated solely from the original textual instruction, semantic ambiguity and spatial uncertainty are directly propagated to the mask layer, compromising its quality and robustness. More critically, for object addition tasks, text-derived masks offer no actionable physical priors (e.g., plausible position, scale, or motion logic for the new object), leaving the model without a grounded basis for generation. Therefore, instruction-based video editing calls for a new framework that can build a robust bridge between high-level semantics and low-level pixels: it should preserve the expressiveness of language while offering fine-grained spatial and physical constraints, while avoiding heavy reliance on large amounts of aligned annotations.

To address this need, we propose a \textit{Plan--Guide--Edit} paradigm for instruction-based video editing, which explicitly structures the mapping from semantic intent to spatial execution. First, a structured planning stage precedes editing, decomposing the originally implicit text comprehension process into interpretable, multi-step reasoning. Specifically, we leverage a Chain-of-Thought (CoT)-enhanced multimodal large language model (MLLM) to perform a hierarchical analysis of input video keyframes and text instructions. It first identifies the editing objects and action intents, and then, while modeling spatial localization, temporal consistency, and physical feasibility, generates a sequence of keyframe bounding boxes to precisely anchor the spatiotemporal positions to be edited. In parallel, it outputs an enriched instruction that supplements the original description with target attributes, interaction tendencies, and scene constraints, providing stronger semantic priors to subsequent modules. Guided by this, mask generation is no longer performed solely from ambiguous text but is instead guided by the explicit spatial prior of the bounding box to generate high-quality local masks, significantly boosting localization accuracy and providing essential constraints on scale and contact relationships, particularly for object addition. Finally, the main diffusion-based editing module integrates the precise spatial constraints from the masks, the enriched semantic control from the augmented instructions, and underlying video features, grounding the high-level semantic intent into specific spatio-temporal locations for more controllable instruction-based video editing.

The key to our proposed framework's ability to resolve the deficiencies of text-only methods lies in its establishment of a reliable semantic--spatial link through structured planning. First, the CoT-enhanced MLLM planner acts as a semantic--spatial translator, proactively resolving textual ambiguities before execution. The explicit generation of bounding boxes decouples the burden of spatial localization from the diffusion model, substantially simplifying subsequent mask generation and editing. Second, our design ensures physical consistency: the planner explicitly reasons about scale, spatial relationships, and kinematics, and the resulting bounding boxes and enriched instructions provide actionable physical priors for the edited objects. Finally, this decoupled \textit{Plan--Guide--Edit} paradigm clarifies the learning objectives of each module and, via staged training, reduces the dependence on large-scale aligned data while improving stability and efficiency.

Our contributions can be summarized as follows:
\begin{itemize}
\item We introduce a structured \textit{Plan--Guide--Edit} paradigm for instruction-based video editing that parses editing targets and localizes them in space and along the motion dimension before editing, enabling a clear mapping from high-level instructions to concrete video edits.
\item We develop a CoT-enhanced MLLM planner that provides structured, physically aware guidance from video and instructions, and introduce a box-conditioned mask prediction branch that converts this guidance into explicit spatial priors to ease the diffusion-based editor's generation.
\item Extensive experiments show that our framework achieves accurate spatial localization and consistent physical and temporal behavior, outperforming existing baselines on complex instruction-based video editing tasks.
\end{itemize}

\begin{figure*}[t]
  \centering
  \includegraphics[width=\linewidth]{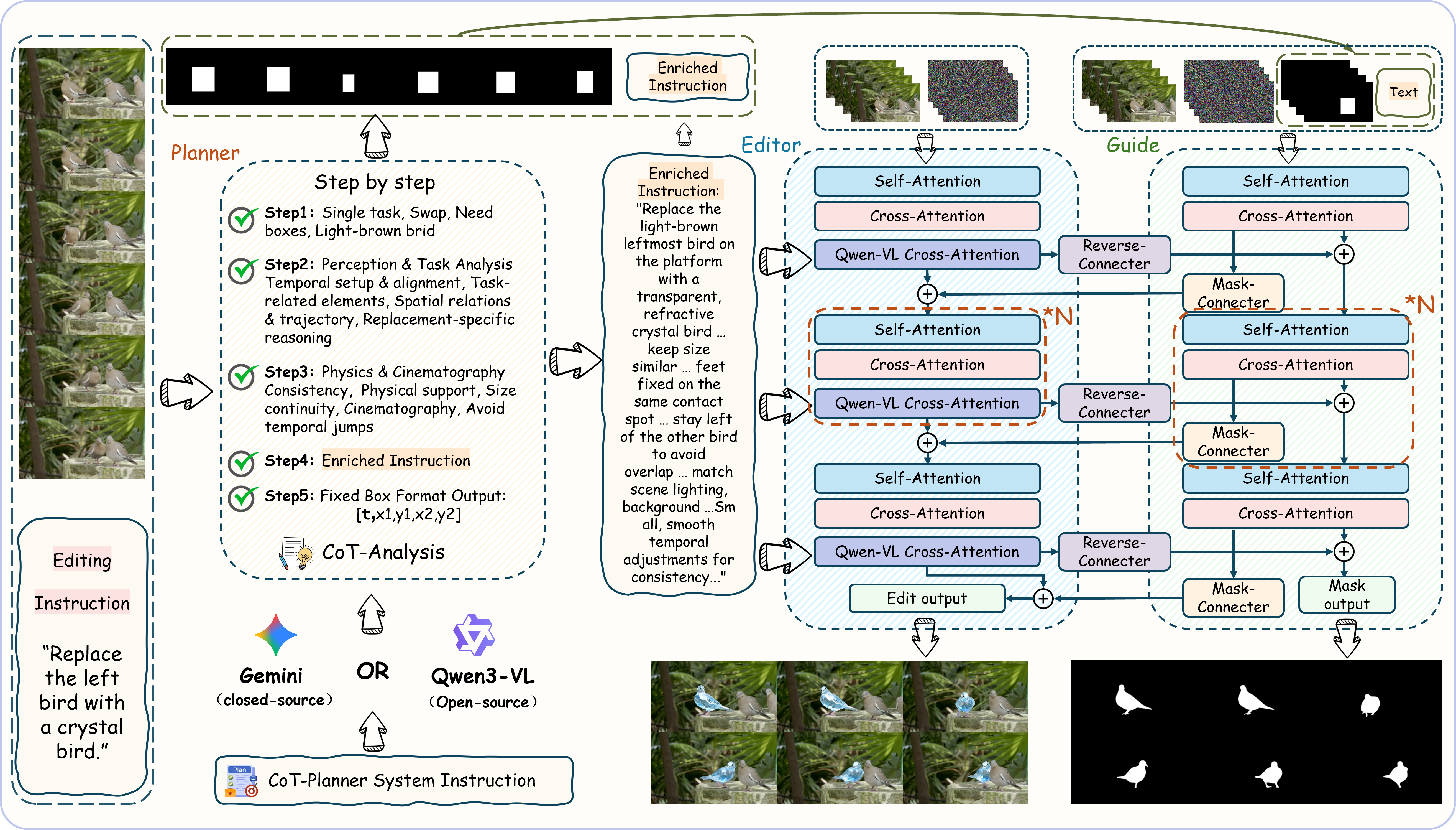}
  \caption{
    Overview of the proposed ``Plan--Guide--Edit'' framework for instruction-based video editing. 
    Given a video and an instruction, a CoT-enhanced VLM planner performs step-by-step analysis to produce an enriched instruction and a temporal sequence of bounding boxes. 
    The Guide branch turns these spatial priors into spatio-temporally consistent masks, while the Editor fuses text, video features, and mask guidance through bidirectional connectors to render the final edited video.
    }

  \label{fig:Method}
\end{figure*}

\section{Related Work}
\label{sec:Related Work}

\subsection{Video Generation}
Early video generation relied on GANs/VAEs but faced low resolution, flickering, and temporal incoherence~\cite{Stylegan-v, wang2025tiv, ding2025rass, gao2025compress, liu2024grounding, hu2025hunyuancustom}. Diffusion models revolutionized the field by extending image diffusion techniques to video, achieving high-fidelity results through temporal denoising and cascaded architectures~\cite{VDM, Imagenvideo, Make-a-video, Phenaki}. Scaling these models enabled large-scale text-to-video systems with broad scene coverage~\cite{Cogvideo, ModelScope, liu2026health}. To enhance control beyond text, image-guided methods emerged, animating static inputs or adapting image models zero-shot~\cite{Animate_anyone, I2vgen-xl, Text2video-zero}. Multi-condition frameworks~\cite{Hunyuanvideo, Vace, ma2025magicstick, chen2025hunyuanvideo, liang2025omniv2v, zhang2025motionpro, ma2024follow} now integrate text, images, poses, masks, and videos to improve controllability, though consistency challenges remain.

\subsection{Instruction Video Editing}
Non-instructional video editing methods~\cite{Tune-A-Video, Video-P2P, wang2026training, wang2025re} rely on auxiliary inputs or per-video fine-tuning, which compromises flexibility and efficiency. To reduce interaction costs, the instruction-guided paradigm emerged, inspired by InstructPix2Pix~\cite{InstructPix2Pix}, which requires only the original video and a natural language instruction. However, it faces dual challenges: the scarcity of large-scale, high-quality instruction-video aligned data, and  maintaining spatio-temporal consistency and physical plausibility during editing. To address the data bottleneck, InstructVid2Vid~\cite{InstructVid2Vid} constructs edit pairs via per-video optimization but suffers from poor scalability. EffiVED~\cite{EffiVED} and InsV2V~\cite{InsV2V} extend image-editing techniques to synthesize pseudo video data, while recent InsViE~\cite{Insvie-1m} and Ditto~\cite{Ditto} scale datasets to millions of samples through model distillation or image-to-video generation pipelines, enabling diverse editing tasks. End-to-end models (e.g., Lucy-1.1~\cite{Lucyedit}) validate the feasibility of video-instruction inputs yet exhibit inadequate spatial localization for small targets or ambiguous instructions. InstructX~\cite{InstructX} enhances semantic generalization using multimodal large models but relies on implicit attention for spatial constraints. In contrast, methods~\cite{StableV2V,Veggie} emphasizing geometric consistency  improve cross-frame coherence via explicit alignment yet require external modules and are not inherently instruction-driven.
However, most instruction-driven methods still struggle to turn ambiguous high-level instructions into precise, physically plausible spatiotemporal edits, underscoring the need for a framework that jointly performs instruction parsing, spatial grounding, and fine-grained editing.

\section{Method}
We propose a \textit{Plan--Guide--Edit} framework for instruction-based video editing tasks, as illustrated in Fig.~\ref{fig:Method}. The framework is implemented through three core modules: the \textbf{Planner}, which translates high-level semantic intent into executable spatial constraints; the \textbf{Guide}, which generates spatiotemporally consistent masks based on explicit spatial priors; and the \textbf{Editor}, which executes appearance modification and content generation. This modular design ensures semantic coherence, spatial precision, and temporal continuity while reducing the complexity of editing tasks.

\noindent 


\begin{figure*}[t]
  \centering
  \includegraphics[width=\linewidth]{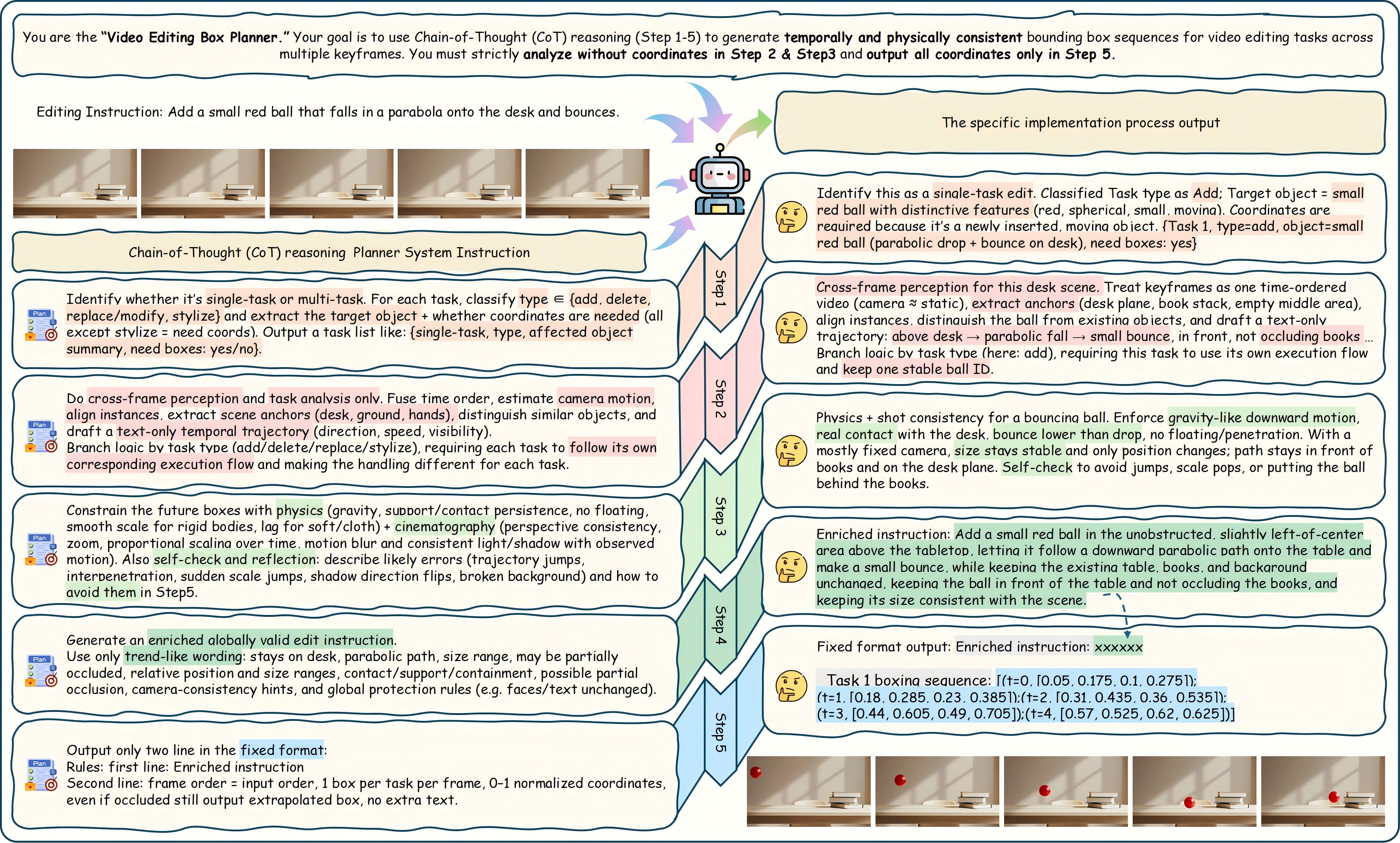}
  \caption{
    Chain-of-Thought (CoT) reasoning process of the MLLM planner. Given an editing instruction and keyframes, the planner follows a five-step procedure: it parses the task type and target object, performs cross-frame perception and temporal analysis, enforces physics- and cinematography-aware consistency, synthesizes an enriched instruction, and finally outputs the enriched instruction together with a temporally coherent sequence of normalized bounding boxes in a fixed format.
  }
  \label{fig:CoT_final}
\end{figure*}

\textbf{Base Model and Setup.} For the latter two branches of our framework, we select Wan2.2 5B as the base diffusion backbone. Given the original video $S$, we obtain its low-dimensional latent representation $y_0$ via a 3D VAE. During the diffusion process, noise is injected to obtain $y_{\text{noise}}$, and the instruction text is denoted as $I$. We concatenate $y_0$ and $y_{\text{noise}}$ along the channel dimension:
\begin{equation}
y_{\text{input}} = \text{ChannelConcat}(y_{\text{noise}}, y_0),
\label{eq:y_input}
\end{equation}
and expand the input channel dimension of the tokenizer $T$ to 32 while keeping the output dimension unchanged, thus avoiding disruption to the original Wan2.2 5B distribution.

\subsection{Planner: Semantic--Spatial Planning with CoT-Enhanced Reasoning}
\label{sec:CoT}
The planner serves the pivotal role of translating high-level semantic intents into executable spatial constraints. We implement it as an MLLM explicitly embedded with a Chain-of-Thought (CoT) reasoning process. This design bridges ambiguous semantic instructions and precise spatial execution before mask prediction and diffusion-based editing, explicitly determining spatial localization, positional relationships, and physical consistency so that downstream modules focus on localized refinement and appearance generation under well-defined spatial priors.

\textbf{Inputs and Outputs.} The planner takes as input a temporally ordered sequence of keyframes $\{I_t\}_{t=1}^T$, together with the user's editing instruction. It produces two parallel, structured outputs: (1) a keyframe-aligned sequence of bounding boxes $\{b_t\}_{t=1}^T$ that explicitly specifies where the edit should occur; and (2) an enriched instruction that extends the original text with semantic priors such as target attributes, relative spatial relations, contact patterns, and camera-consistency hints, which are used to condition the mask and generation branches. For non-spatial tasks (e.g., stylization), the planner emits an empty box sequence.

\vspace{-5pt}
\subsubsection*{Structured Reasoning Procedure}
As illustrated in Fig.~\ref{fig:CoT_final}, we organize the planner's reasoning into an explicit multi-step Chain-of-Thought process instead of a direct single-pass output. This is motivated by three factors: (i) sequentially decomposing complex instructions significantly reduces task difficulty; (ii) pre-modeling physical and cinematic constraints prevents localization errors from amplifying in later generation stages; and (iii) a dedicated stage creates structured, interpretable guidance for downstream modules. Based on this, we decompose the planning process into three tightly connected components described below.
\begin{enumerate}
    \item[(i)] \textbf{Task Parsing and Cross-Frame Perception.}
    The objective of this stage is to establish an accurate understanding of both the editing intent and the video content. On the instruction side, the planner parses the text to determine the task type (e.g., addition, swap) and extracts the primary objects. On the visual side, it performs cross-frame perception: estimating camera motion (to distinguish from object motion) and conducting cross-frame task-relevant instance identification and tracking. This stage thus anchors an initially ambiguous linguistic reference to concrete video instances, providing a reliable basis for subsequent precise localization and temporal reasoning.

    \item[(ii)] \textbf{Physics and Temporal Consistency Modeling.}
    This stage is a critical component of our planner, compelling the MLLM to explicitly model physical constraints and cross-frame continuity. First, inter-frame consistency reasoning requires the MLLM to infer the target's displacement, scale changes, and visibility across frames to generate a smooth localization trajectory by integrating object motion, camera motion, and perspective changes. Second, physical rule embedding leverages the MLLM's world knowledge for constraints (e.g., gravity, contact, occlusion), which, combined with a final self-check and reflection step, mitigates common failure modes in purely text-driven editing, such as inaccurate localization or physical implausibility.

    \item[(iii)] \textbf{Generating Spatial and Semantic Guidance.}
    Following rigorous parsing and consistency modeling, the planner produces its two structured outputs:
    \begin{itemize}
        \item \textbf{Temporal Bounding Box Sequence.} A sequence of normalized bounding boxes $[t, x_1, y_1, x_2, y_2]$, denoted $\{b_t\}_{t=1}^T$, is generated, aligned with all keyframes. This sequence specifies \textit{where to edit} and ensures plausible and coherent locations.
        \item \textbf{Structured Enriched Instruction.} The planner further produces an augmented textual instruction that preserves the original user intent while injecting attributes discovered during reasoning, such as key spatial relations, likely interactions, motions, and scene-level constraints.
    \end{itemize}
    This dual-channel guidance decouples rigid spatial constraints (boxes) from flexible semantic control (enriched instruction): the former provides verifiable spatial anchors and the latter supplies transferable semantic and physical priors, forming a robust prior for mask generation and diffusion editing.
\end{enumerate}

\subsection{Guide: Box-Guided Mask Generation with Spatiotemporal Constraints}

Given the planner's outputs, namely the bounding box sequence $B = \{b_t\}_{t=1}^T$ and the enriched instruction $EI$, we transform mask generation from a difficult global problem into a local refinement problem under explicit positional priors. Concretely, this branch treats $B$ as hard constraints and uses video features together with $EI$ as soft constraints to generate a binary mask $M$. The resulting spatiotemporal features at layer $l$, denoted $C_l^M$, are used as implicit spatial guidance for the downstream editing module. Leveraging the accurate spatial anchors provided by the planner, the mask predictor no longer needs to solve a hard localization-and-generation problem, and instead only needs to infer the precise shape required by the editing operation within the specified regions.

\textbf{Spatiotemporal self-attention and cross-modal constraints.}
The mask prediction branch (i.e., the Guide) shares the same hierarchical architecture as the Editor. Self-attention is propagated temporally and across neighboring scales. Cross-attention connects to two types of conditions: global video features that compensate for context, and the planner's outputs ($EI$ and $B$) that inject semantic constraints and accurate spatial position priors. This design greatly reduces the difficulty of mask prediction, allowing the generated masks to correctly capture the true spatial footprint of the target and its motion or contact relationships, thereby yielding semantically meaningful mask outputs.

\vspace{5pt}
\textbf{Reverse-Connector to Editor Branch.} 
Masks generated solely from bounding boxes and local features can be unstable for thin structures or heavily occluded regions. To address this, we introduce a Reverse-Connector from the Editor back to the Guide:
\begin{equation}
C_l^M = C_l^M + \text{Reverse\_Connector}(Q_l^E),
\label{eq:reverse_connector}
\end{equation}
where $Q_l^E$ denotes the editor features at layer $l$. This operation performs a backward correction based on the editor's understanding. Intuitively, the semantic understanding of the editing requirements helps the mask branch recover details that are otherwise easy to miss, and the connection is multi-layer and repeatable.

\vspace{5pt}
\textbf{Mask-Connector and Explicit Mask Output.}
For the Guide branch, we use a \emph{Mask-Connector} to map the refined spatiotemporal features to mask maps that share the same spatial resolution as the input frames. Concretely, we apply a per-frame decoding head, enforce constraints from the bounding boxes, preserve the temporal order to maintain inter-frame continuity of the masks, and finally output a set of masks $\{M_t\}$ that can be directly supervised. These masks can be used as explicit spatial conditions for diffusion-based editing and can also participate in joint optimization during training, stabilizing the learning of the backbone.

In parallel, we inject the mask information back into the Editor branch using
\begin{equation}
Q_l^E = Q_l^E + \text{Mask\_Connector}(C_l^M),
\label{eq:mask_connector}
\end{equation}
where $\text{Mask\_Connector}(\cdot)$ projects the mask features $C_l^M$ at layer $l$ to an additive modulation of the editor features $Q_l^E$. By explicitly injecting the current-frame mask into several layers, the editor features at different depths remain aware of low-level spatial guidance.

With these design choices, the mask branch maps instructions to temporally consistent masks: the planner resolves semantic ambiguities, bounding boxes anchor spatial uncertainty, and spatiotemporal mechanisms mitigate temporal instability, thereby providing an accurate guiding condition for the editing branch.

\subsection{Editor: Diffusion-Based Editing with Multi-Condition Fusion}
Given the precise spatiotemporal masks $\{M_t\}$ and the structured enriched instruction $EI$, the Editor branch performs appearance modification and content generation. Its goal is to inject these conditions into the diffusion generator, preserving unedited regions and producing visually natural, plausible, and smooth results.

\vspace{5pt}
\textbf{Conditioned diffusion backbone.}
In this branch, cross-attention modules receive the enriched instruction $EI$. In particular, we take the last-layer visual-language features $V$ from Qwen-VL and inject them into a dedicated Qwen-VL cross-attention module. We keep only the first 1024 tokens of $V$ and map their channel dimension from 3584 down to the 3072 channels used by Wan2.2 5B through a multilayer perceptron (MLP).
We then augment each editing block with a specialized cross-attention to these MLLM features:
\begin{equation}
Q_l^E = Q_l^E + \text{QVLcrossattn}(\text{MLP}(V), C_l^M),
\label{eq:qvl_cross_attn}
\end{equation}
where $\text{QVLcrossattn}(\cdot,\cdot)$ denotes cross-attention from the editor features $Q_l^E$ to the projected MLLM tokens, modulated by the mask features $C_l^M$. In this way, the Editor not only receives visual priors from the original video but also fully exploits the semantic and world knowledge that the MLLM has already organized in the planning stage.

\vspace{5pt}
\textbf{Bidirectional collaboration with the mask branch.}
The masks are not fed into the Editor only once; instead, the Editor and Guide cooperate bidirectionally through multi-layer interactions. As described in the previous section, the Reverse-Connector feeds high-level semantics from the Editor back to the mask branch, while the Mask-Connector injects the mask into multiple editor layers, ensuring multi-depth features remain aware of this low-level guidance. This bidirectional coupling improves the stability of the generated results in terms of boundaries, occlusions, and lighting, which is especially beneficial for fine-grained local edits.

\vspace{-5pt}
\paragraph{Framework Loop.}
In the \textit{Plan--Guide--Edit} framework, the Planner first parses ambiguous language into ordered boxes and an enriched instruction. Anchored by these boxes, the Guide produces spatiotemporal masks, which it refines using feedback from the Editor. The Editor then injects both masks and instructions into the diffusion backbone to perform conditional generation. Finally, a mask-guided compositing strategy applies edited content inside the mask regions while preserving the original video outside, ensuring a temporally coherent, semantically faithful, and physically compatible output.

\begin{figure*}[t]
  \centering
  \includegraphics[width=\linewidth]{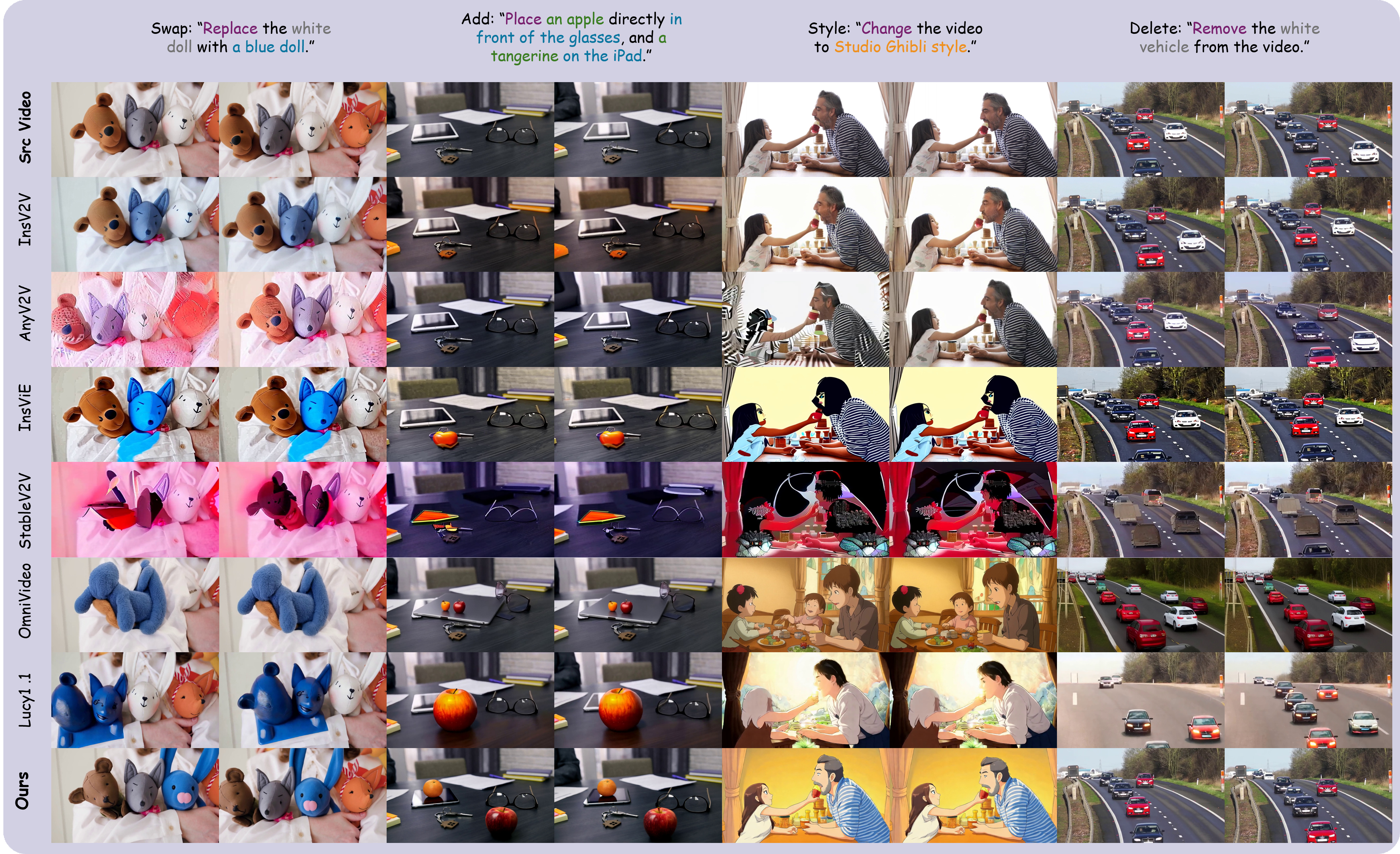}
  \caption{
    Qualitative comparison of CoT-Edit and open-source baselines, showing more precise target localization, better preservation of non-edited regions, and more faithful handling of complex, physics-aware editing instructions.
  }
  \label{fig:Experiment}
\end{figure*}

\section{Experiments}

\subsection{Experimental Settings}

\noindent\textbf{Implementation details.}
We adopt a two-stage training strategy. In the first stage, the Editor branch and the Mask (Guide) branch are trained separately.  
The \textit{Mask branch} is trained on a mixture of image segmentation datasets (ADE20K~\cite{ADE20k}, PhraseCut~\cite{PhraseCut}) and video segmentation datasets (YouTube-VOS~\cite{Refer-Youtube-VOS}, OVIS~\cite{refOVIS}).  
The \textit{Editor branch} is trained on two types of data: (i) open-source image/video editing datasets AnyEdit~\cite{Anyedit}, UltraEdit~\cite{Ultraedit}, SEED-Data-Edit~\cite{Seed-data-edit}, EditWorld~\cite{Editworld}; (ii) open-source video instruction editing datasets Señorita-2M~\cite{zi2025se}, Ditto~\cite{Ditto}, together with our own instruction-based video editing dataset.

In the second stage, we jointly train the two branches on our internal dataset, which contains about 100k high-quality video editing pairs (before/after) with precise mask annotations. We train the model at a resolution of $720 \times 1280$. The first stage runs for 20k steps and the second stage for 10k steps, with batch size 64 for both stages. The Reverse-Connector and Mask-Connector are initialized with zeros.

\noindent\textbf{Evaluation metrics.}
We evaluate our method from two perspectives: overall visual quality of the edited videos and instruction-following capability.  
For visual quality, we use FVD~\cite{FVD} and the VBench~\cite{Vbench} metrics Background Consistency (BC), Temporal Consistency (TC), Motion Smoothness (MS), and Aesthetics (AES).  
For instruction following, we employ CLIPScore~\cite{Clipscore} and use the Gemini~\cite{Gemini} model to rate the edited videos in terms of physical plausibility, spatial relations, instruction following, and overall editing quality.

\begin{itemize}
    \item \textbf{Background Consistency (BC).} We compute cross-frame similarity in the CLIP feature space to measure temporal consistency of the global background.
    \item \textbf{Temporal Consistency (TC).} Following VBench, we use CLIP-B to compute similarities between each frame and its adjacent frames, as well as the first frame, to assess temporal coherence.
    \item \textbf{Motion Smoothness (MS).} Motion smoothness measures the continuity and naturalness of subject or camera motion. A normal video should avoid jitter and unnatural acceleration changes. We adopt motion priors from a video frame interpolation model to evaluate motion smoothness of edited videos.
    \item \textbf{Aesthetics (AES).} We use the LAION aesthetic predictor to score the aesthetic quality of each frame.
    \item \textbf{CLIPScore.} We use CLIP-B to measure the alignment between the text instruction and the generated video.
\end{itemize}

\noindent\textbf{Evaluation protocol.}
We randomly sample 100 videos from the Koala-36M~\cite{Koala-36m} dataset and assign diverse editing instructions, including object addition, deletion, attribute modification, stylization, and physically plausible motions such as parabolic trajectories.

\begin{table*}[ht]
\centering
\renewcommand{\arraystretch}{1.1}
\setlength{\tabcolsep}{4pt} 
\caption{
Quantitative comparison of CoT-Edit and baseline models on visual and editing quality metrics.
}
\begin{tabular}{l|cccccc|cccc}
\hline
MODEL &
BC\,{$\uparrow$} &
CLIPScore\,{$\uparrow$} &
FVD\,{$\downarrow$} &
TC\,{$\uparrow$} &
MS\,{$\uparrow$} &
AES\,{$\uparrow$} &
\begin{tabular}{c}
Physical\\
Rule\,{$\uparrow$}
\end{tabular} &
\begin{tabular}{c}
Spatial\\
Relation\,{$\uparrow$}
\end{tabular} &
\begin{tabular}{c}
Instruction\\
Following\,{$\uparrow$}
\end{tabular} &
\begin{tabular}{c}
Editing\\
Quality\,{$\uparrow$}
\end{tabular} \\
\hline
InsV2V    & 0.921 & 0.129 & 4095.42 & 0.958 & 0.95 & 0.57 & 0.367 & 0.291 & 0.401 & 0.326 \\
StableV2V & 0.942 & 0.263 & 3222.18 & 0.925 & 0.99 & 0.46 & 0.305 & 0.341 & 0.381 & 0.319 \\
InsViE    & 0.936 & 0.395 & 2397.65 & 0.957 & 1.10 & 0.48 & 0.389 & 0.285 & 0.355 & 0.369 \\
AnyV2V    & 0.927 & 0.251 & 2942.77 & 0.918 & 0.95 & 0.41 & 0.313 & 0.373 & 0.265 & 0.394 \\
OmniVideo & 0.972 & 0.382 & 1076.44 & 0.954 & 1.04 & 0.44 & 0.590 & 0.641 & 0.526 & 0.604 \\
Lucy-1.1  & 0.931 & 0.376 & 1488.12 & \textbf{0.961} & 0.98 & 0.52 & 0.488 & 0.769 & 0.562 & 0.589 \\
Ours      & \textbf{0.974} & \textbf{0.445} & \textbf{1015.67} & 0.945 & \textbf{1.18} & \textbf{0.62} & \textbf{0.741} & \textbf{0.841} & \textbf{0.629} & \textbf{0.648} \\
\hline
\end{tabular}

\label{table:Experiment}
\end{table*}

\begin{table}[t]
\centering
\caption{
Ablation study results on video editing. Higher is better. PR, SR, IF, and OEQ denote physical plausibility, spatial relations, instruction following, and overall editing quality, respectively.
}
\begin{tabular}{lcccc}
\toprule
 & PR & SR & IF & OEQ \\
\midrule
E w/o MLLM           & 0.501 & 0.754 & 0.575 & 0.598 \\
E w/ MLLM            & 0.674 & 0.758 & 0.598 & 0.631 \\
E + M w/ Mc          & 0.643 & 0.807 & 0.586 & 0.638 \\
E + M w/ Mc \& Rc    & 0.681 & 0.815 & 0.609 & 0.647 \\
\bottomrule
\end{tabular}

\label{table:ablation}
\end{table}

\subsection{Comparison with State-of-the-Art Methods}

\noindent\textbf{Baselines.}
We compare our proposed CoT-Edit with several state-of-the-art open-source instruction-based video editing methods, including InsV2V~\cite{InsV2V}, StableV2V~\cite{StableV2V}, InsViE~\cite{Insvie-1m}, AnyV2V~\cite{Anyv2v}, OmniVideo~\cite{Omni-video}, and Lucy-1.1~\cite{Lucyedit}. OmniVideo~\cite{Omni-video} combines MLLMs with diffusion models in a unified framework, while InsViE~\cite{Insvie-1m} and Lucy-1.1~\cite{Lucyedit} perform end-to-end video editing purely with diffusion models. However, due to their relatively weak instruction injection mechanisms, these methods are limited in modeling physical laws, spatial relations, instruction following, and overall editing quality.

\noindent\textbf{Qualitative results.}
We present qualitative comparisons between CoT-Edit and other open-source methods in Fig.~\ref{fig:Experiment}. CoT-Edit demonstrates:  
(1) precise localization and editing of the target object specified in the instruction among multiple similar objects;  
(2) strong preservation of non-edited regions consistent with the original video;  
(3) robust understanding of complex editing instructions and high-quality multi-task editing.  
Moreover, CoT-Edit supports instruction-based video edits that explicitly follow physical laws, such as parabolic motion or uniform linear motion. More results are shown in the supplementary material.

\noindent\textbf{Quantitative results.}
We quantitatively compare CoT-Edit with baseline models using multiple automatic metrics, summarized in Table~\ref{table:Experiment}. For visual quality, we report FVD~\cite{FVD} and VBench~\cite{Vbench} metrics (BC, TC, MS, AES). For editing quality, we rely on Gemini~\cite{Gemini} to score physical plausibility, spatial relations, instruction following, and overall editing quality of the edited videos. CoT-Edit consistently outperforms all baselines across most dimensions, with particularly large gains in spatial reasoning and physical law understanding.

\subsection{Further Analysis and Ablation Studies}

\begin{figure}[t]
  \centering
  \includegraphics[width=\linewidth]{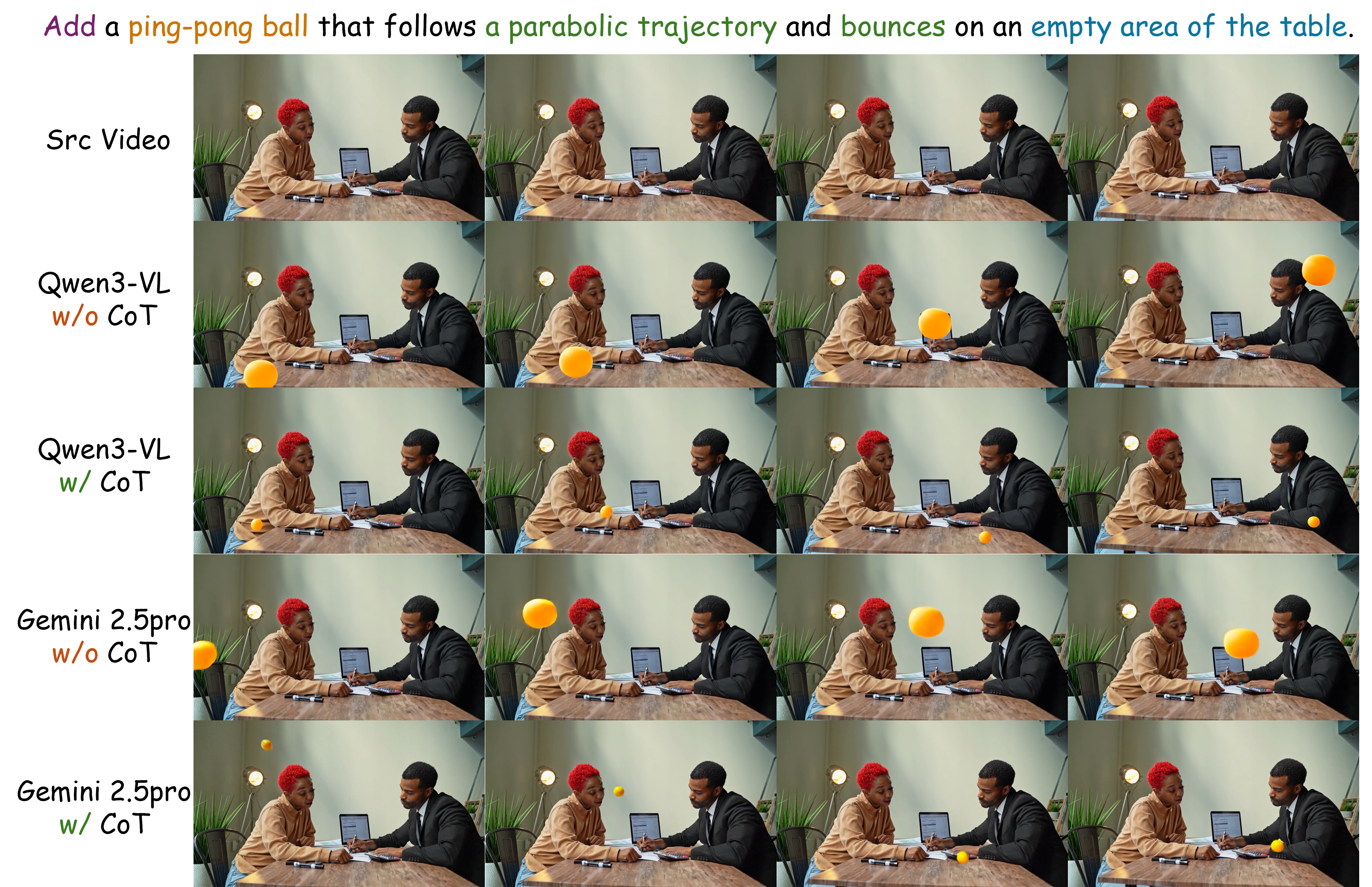}
  \caption{
    Qualitative results of the CoT ablation study.
  }
  \label{fig:Ablation}
\end{figure}

\noindent\textbf{CoT ablation.}
In Section~\ref{sec:CoT}, we introduce the CoT scheme, which decomposes instruction understanding into intermediate reasoning steps and outputs both the editing region locations and an enhanced instruction text. To verify its effectiveness, we compare four variants: (i) Qwen3-VL-32B; (ii) Qwen3-VL-32B with CoT; (iii) Gemini 2.5 Pro~\cite{Gemini}; (iv) Gemini 2.5 Pro with CoT.  
Qualitative results in Fig.~\ref{fig:Ablation} show that, without CoT, the ping-pong ball fails to exhibit physically plausible motion. In contrast, with CoT, both variants generate a clear parabolic motion where the ball first follows a parabolic trajectory, hits the table, and then bounces upwards, closely following physical laws. These results indicate that our CoT scheme not only improves target object localization but also enables physically consistent motion generation.

\noindent\textbf{Mask branch ablation.}
The proposed Mask branch is responsible for task decomposition and region localization. It provides implicit mask signals to guide the Editor branch on where to edit, thus reducing the learning burden of the Editor. In the Editor branch, we further introduce Qwen3-VL-8B as an instruction enhancement module: the original video and instruction are first processed by Qwen3-VL, and the extracted features are injected into the Editor to improve instruction understanding.  
We compare four configurations corresponding to those in Table~\ref{table:ablation}: (i) E w/o MLLM, the Editor branch without Qwen3-VL; (ii) E w/ MLLM, the Editor branch with Qwen3-VL features; (iii) E + M w/ Mc, the Editor plus the Mask branch with Mask-Connector; (iv) E + M w/ Mc \& Rc, the full model with both Mask-Connector and Reverse-Connector. Experimental results in Table~\ref{table:ablation} show that Qwen3-VL significantly improves instruction-following performance, while the Mask branch enhances the Editor's understanding of spatial locations.

\noindent\textbf{User study.}
To complement automatic metrics, we conducted a user study assessing perceptual quality and physical consistency. Participants ranked CoT-Edit against baseline methods on visual quality and physical consistency. The results showed a clear preference for CoT-Edit in all aspects. Detailed protocols and statistics are provided in the supplementary material.

\section{Conclusion}



We presented a \textit{Plan--Guide--Edit} framework where a CoT-enhanced MLLM planner, a box-guided mask branch, and a diffusion editor link semantic intent to spatial execution. By explicitly modeling spatial constraints and priors, our method achieves precise localization, temporal coherence, and preservation of unedited regions. Experiments show consistent gains over prior text-driven methods in spatial reasoning and visual quality.

\clearpage
\section*{Acknowledgments}
This work was supported in part by NSFC under Grant 62371434, the Postdoctoral Fellowship Program of CPSF under Grant Number GZC20252293, the China Postdoctoral Science Foundation-Anhui Joint Support Program under Grant Number 2024T017AH, China Postdoctoral Science Foundation under Grant Number  2025M783529, Anhui Postdoctoral Scientific Research Program Foundation (No.2025A1015), the Fundamental Research Funds for the Central Universities(No. WK2100250064), and ZGCA Project-C20250302.

{
    \small
    \bibliographystyle{ieeenat_fullname}
    \bibliography{main}
}


\end{document}